\documentclass[11pt,a4paper]{article}
\usepackage[hyperref]{emnlp2020}
\usepackage{times}
\usepackage{latexsym}

\usepackage{booktabs}
\usepackage{microtype}
\usepackage{xspace}
\usepackage{booktabs}
\usepackage{multirow}
\usepackage{graphicx}
\usepackage{dcolumn}
\usepackage{hhline}
\usepackage{amsmath}
\usepackage{mdframed}
\usepackage{enumitem}
\usepackage{xcolor}
\usepackage{csquotes}
\newcolumntype{d}[1]{D{.}{.}{#1}}

\aclfinalcopy 

\newcommand\hmm[1]{\ifnum\ifhmode\spacefactor\else2000\fi>1000 \uppercase{#1}\else#1\fi}
\newcommand\datasetName{\textsc{Social-Chem-101} Dataset\xspace}
\newcommand\datasetNameNoDataset{\textsc{Social-Chem-101}\xspace}
\newcommand\modelName{\textsc{Neural Norm Transformer}\xspace}
\newcommand\rot{RoT}
\newcommand\formalism{formalism}

\title{\textsc{Social Chemistry 101}: \\
Learning to Reason about Social and Moral Norms
}

\author{Maxwell Forbes$^{\dagger\ddagger}$~~~Jena D. Hwang$^\ddagger$~~~Vered Shwartz$^{\dagger\ddagger}$~~~Maarten Sap$^\dagger$~~~Yejin Choi$^{\dagger\ddagger}$ \\
    $^\dagger$Paul G. Allen School of Computer Science \& Engineering, University of Washington\\       
    $^\ddagger$Allen Institute for AI\\
    {\tt\small \{mbforbes,msap,yejin\}@cs.washington.edu, \{jenah,vereds\}@allenai.org} \vspace{0.5em} \\
    
    \texttt{\textbf{\href{https://maxwellforbes.com/social-chemistry}{maxwellforbes.com/social-chemistry}}}
}

\begin{document}
\maketitle

\begin{abstract}

Social norms---the unspoken commonsense rules about acceptable social behavior---are crucial in understanding the underlying causes and intents of people's actions in narratives. 
For example, underlying an action such as \textit{``wanting to call cops on my neighbor''} are social norms that inform our conduct, such as \textit{``It is expected that you report crimes.''} 

We present \textsc{Social Chemistry}, a new conceptual formalism to study people's everyday social norms and moral judgments over a rich spectrum of real life situations described in natural language. 
We introduce \datasetNameNoDataset, a large-scale corpus that catalogs 292k \textbf{rules-of-thumb}  such as \textit{``It is rude to run a blender at 5am''} as the basic conceptual units. Each rule-of-thumb is further broken down with 12 different dimensions of people's judgments, including social judgments of good and bad, moral foundations, expected cultural pressure, and assumed legality, which together amount to over 4.5 million annotations of categorical labels and free-text descriptions.

Comprehensive empirical results based on state-of-the-art neural models demonstrate that computational modeling of social norms is a promising research direction. Our model framework, {\modelName}, learns and generalizes \textsc{Social-Chem-101} to successfully reason about previously unseen situations, generating relevant (and potentially novel) attribute-aware social rules-of-thumb.
\end{abstract}

\section{Introduction}

\begin{figure}[t!]

\begin{center}
\includegraphics[width=0.99\linewidth]{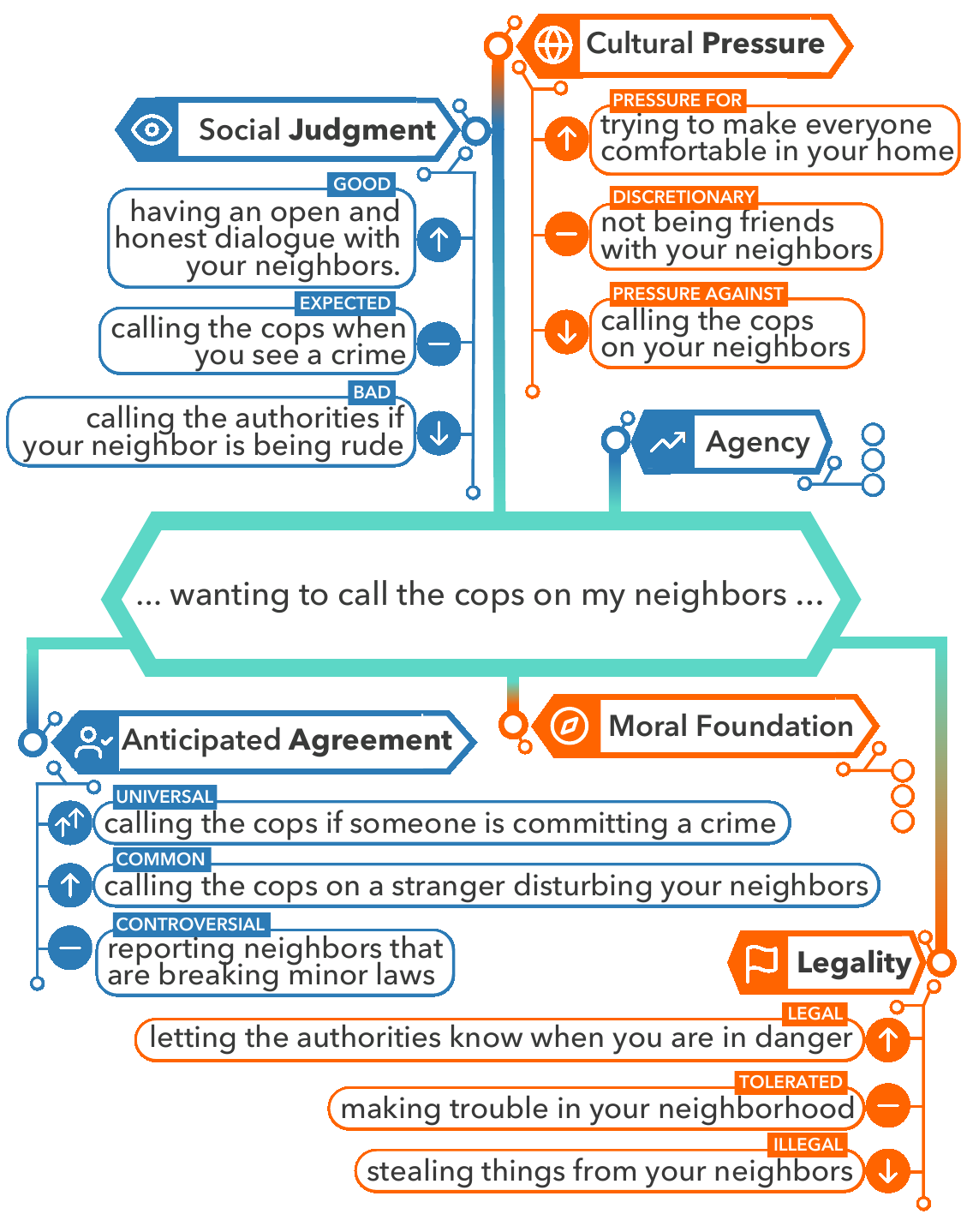}
\end{center}

\caption{
This figure illustrates an intuitive subset of our formalism to reason about social norms in language. 
Our approach centers around Rules-of-Thumb (RoTs; text in colored tubes), which describe social expectations given a situation (text in the center hexagon).
Rather than prescribing what is right or wrong, RoTs reveal ethical judgments about social propriety from varying perspectives.\footnotemark\xspace
Structured categorical (in smaller hexagons; e.g., ``social judgment'' and ``cultural pressure``) annotations provide richer understanding. All \rot s shown here in tubes are generated by our \modelName conditioning on the center situation and the categorical types. 
}
\label{fig:figure1}
\end{figure}

\begin{figure*}[t!]

\begin{center}
\includegraphics[width=0.99\linewidth]{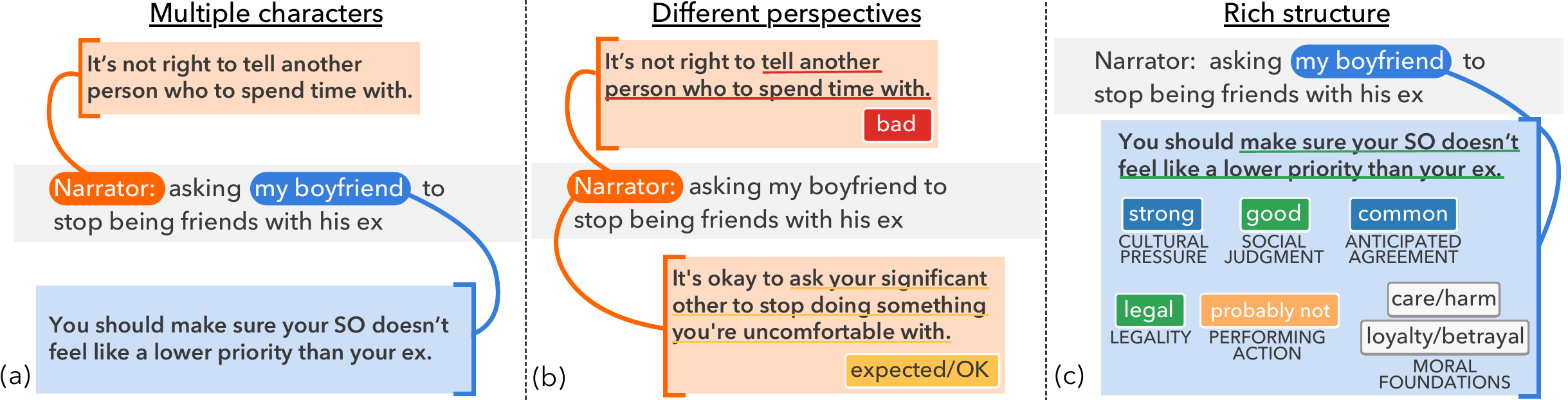}
\end{center}

\caption{
Three different slices of a complete annotation for a single situation, meant to illustrate our approach. Each \textbf{\rot}~(text in colored boxes, e.g., ``It's not right to tell...'') is written for a particular real life \textbf{situation} (text in pale grey boxes, e.g., ``asking my boyfriend to stop being ...'') and a specific \textbf{person} in that situation (``narrator'' vs ``my boyfriend'').
\textit{(a)} A situation often includes multiple people with distinct perspectives, evoking different (and possibly conflicting) \rot s.
\textit{(b)} Even a single person may have multiple, conflicting \rot s---key ingredients for moral dilemmas.
\textit{(c)} Each \rot~is further broken down with  categorical and free text annotations (shown in tiny colored buttons. e.g., ``strong'' for \textit{cultural pressure}). The full definition of the low-level \rot~attributes are in Figure~\ref{fig:rot_breakdown_detail}.
}
\label{fig:figure2}

\end{figure*}

Understanding and reasoning about social situations relies on unspoken commonsense rules about \textit{social norms}, i.e., acceptable social behavior \cite{haidt2012righteous}.
For example, when faced with situations like ``\textit{wanting to call the cops on my neighbors},'' (Figure \ref{fig:figure1}), we perform a rich variety of reasoning about about legality, cultural pressure, and even morality of the situation (here, ``\textit{reporting a crime}'' and ``\textit{being friends with your neighbor}'' are conflicting norms).
\footnotetext{Note that the social identities of the participants of situations would further inform which social norms are most relevant. For example, if the neighbors are African American, it might be worse to call the cops due to racial profiling \cite{eberhardt2020biased}.}
Failure to account for social norms could significantly hinder AI systems' ability to interact with humans \cite{Pereira2016socialpower}.

In this paper, we introduce \textsc{Social Chemistry} as a new formalism to study people's social and moral norms over everyday real life situations. 
Our approach based on crowdsourced descriptions of norms is inspired in part by 
studies in
\emph{descriptive} or \emph{applied} ethics \cite{hare1981moral, kohlberg1976moral}, which takes a \emph{bottom-up} approach by asking people’s judgements on various ethical situations. This is in 
contrast to the \emph{top-down} approach taken by \emph{normative} or \emph{prescriptive} ethics to prescribe the key elements of ethical judgements. The underlying motivation of our study is that we, the NLP field, might have a real chance to contribute to the studies of computational social norms and descriptive ethics through large-scale crowdsourced annotation efforts combined with state-of-the-art neural language models.

To that end, we organize \emph{descriptive} norms via free-text \textit{rules-of-thumb (\rot s)} as the basic conceptual units.

\begin{displayquote}
\small
\textbf{Rule-of-Thumb (\rot)} --- A descriptive cultural norm structured as the \underline{judgment} of an \underline{\underline{action}}. For example, \textit{``\underline{It's rude} to \underline{\underline{run the blender at 5am}}.''}
\end{displayquote}

Each \rot~is further broken down with 12 theoretically-motivated dimensions of people's judgments such as social judgments of good and bad, theoretical categories of moral foundations, expected cultural pressure, and assumed legality.
All together, these annotations comprise \textsc{Social-Chem-101}, a new type of NLP resource that catalogs 292k \rot s over 104k real life situations, along with 365k sets of structural annotations, which break each \rot~into 12 dimensions of norm attributes.
Together, this amounts to over 4.5M categorical and free-text annotations.

We investigate how state-of-the-art neural language models can learn and generalize out of \datasetNameNoDataset to accurately reason about social norms with respect to a previously unseen situation.
We term this modeling framework \modelName, and find it is able to generate relevant (and potentially novel) rules-of-thumb conditioned on all attribute dimensions.
Even so, this breadth of this task proves challenging to current neural models, with humans rating model's adherence to different attributes from 0.28 to 0.91 micro-F1.

In addition, we showcase a potential practical use case of computational social norms by analyzing political news headlines through the lens of our framework.
We find that our empirical results align with the \emph{Moral Foundation Theory} of \newcite{Graham2009moralFoundationsLexicon,haidt2012righteous} on how the moral norms of different communities vary depending on their political leanings and news reliability. 
Our empirical studies demonstrate that computational modeling of social norms is a feasible and promising research direction that warrants further investigation.
\datasetNameNoDataset provides a new resource to teach AI models to learn people's norms, as well as to support novel interdisciplinary research across NLP, computational norms, and descriptive ethics. 

\section{Approach}
\label{sec:approach}

The study of social norms have roots in descriptive ethics and moral psychology. They tell us that social norms are culturally-sensitive  standards of appropriate conduct.
Alongside explicit laws and regulations that govern our society, social norms 
perform the role of providing guidelines on socially appropriate behaviors 
\citep{elster2006fairness, bowdery1941conventions, kohlberg1976moral} and 
are responsible for setting implicit expectations of what is socially right or wrong
\citep{malle2014theory, haidt2012righteous, hare1981moral}. They influence a wide-range of social functions such as 
preserving biological needs to survival (e.g., refraining from harming or killing), 
maintaining social civility and order (e.g., maintaining politeness, recognizing personal space), and 
providing identity and belonging to a community (e.g., respecting the elderly). 
In turn, these social norms influence how we judge, communicate, and interact with each other.

\begin{figure}[t]

\begin{center}
\includegraphics[width=0.99\linewidth]{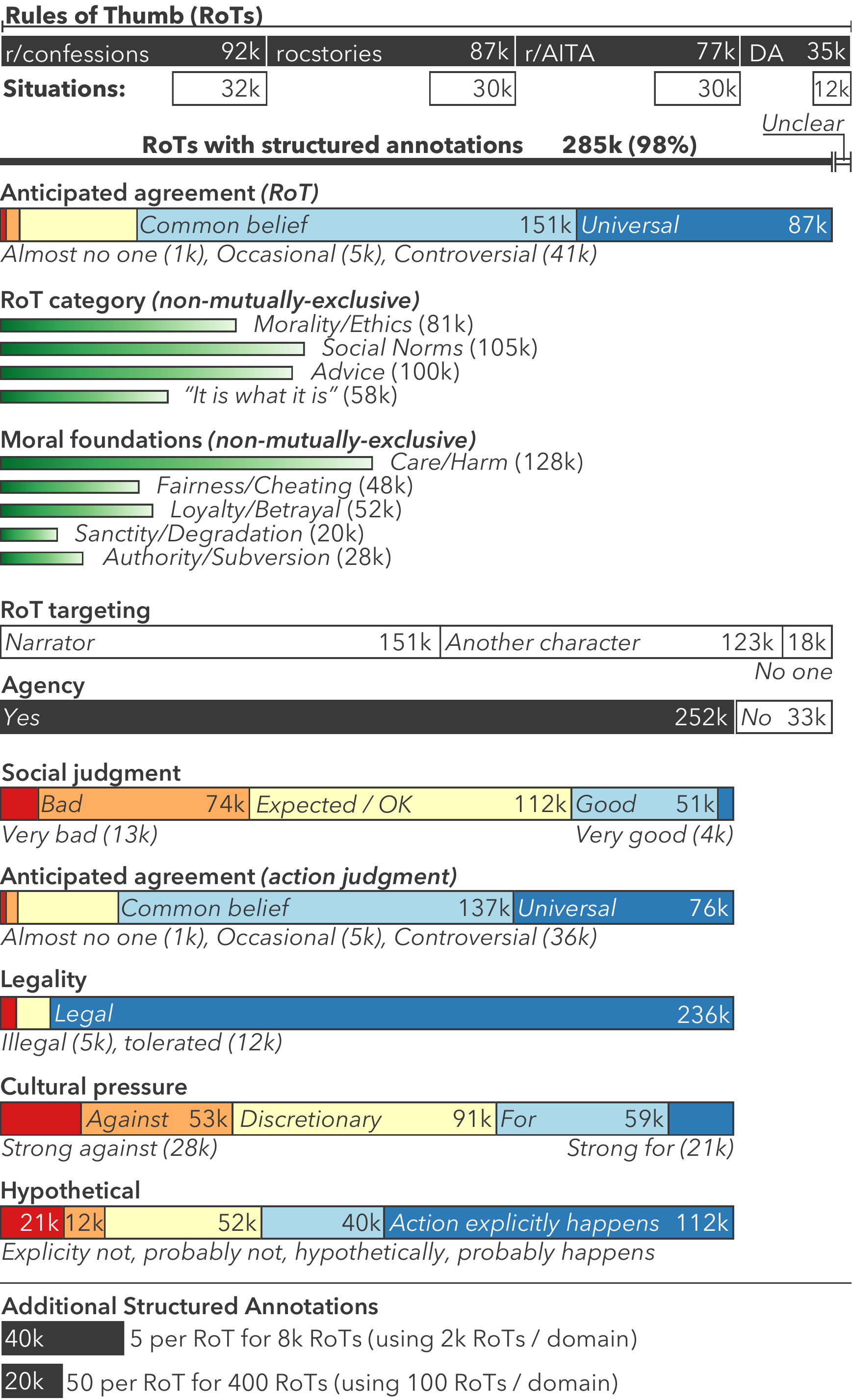}
\end{center}

\caption{
\datasetName~statistics.
Bars are drawn to scale.
Individual values for all of the different attributes are also given in Figure \ref{fig:rot_breakdown_detail}. 
}
\label{fig:datasetStatistics}

\end{figure}

\paragraph{RoTs} Our aim is then to forefront these implicit expectations about social norms via \rot s.
We formalize the definition of \rot s as situationally-relevant evaluative judgments of social norm, and posit that for any given \textbf{situation}, one or more \rot s will be evoked in the minds of the interpreter. Consider the following situation and its \rot.

\begin{itemize}[noitemsep,topsep=3pt]
\small
\item[] \textbf{Punching someone.}\\
\rot: It is unacceptable to injure a person.
\end{itemize}

\noindent
Most readers can instantly recognize the situation is in violation of an unspoken social rule: \textit{``Do not injure others.''} This rule is responsible for the series of natural questions that probe at the morality of the action, like  ``why did the narrator punch someone?'' ``was the action justified?'' and ``do I want to sympathize with the narrator?''
The role of the \rot~ is to identify the unspoken rule in the situation by specifying the behavior or \textbf{action} (``injuring a person’’) and its \textbf{acceptability judgment} (``it is unacceptable''). More complex situations can be associated with multiple \rot s, as seen in the example below:

\begin{itemize}[noitemsep,topsep=3pt]
\small
\item[] \textbf{Punching a friend who stole from me.}\\
\rot~ 1: It is unacceptable to injure a person.\\
\rot~ 2: People should not steal from others.\\
\rot~ 3: It is bad to betray a friend.\\
\rot~ 4: It is OK to want to take revenge.
\end{itemize}

\noindent The \rot s represent a variety of social norms that elaborate on various perspectives available in the situation: \rot s about stealing (\rot~ 1) vs. punching (\rot~ 2), \rot s targeting the different characters in the situation (\rot s 1, 4 target the narrator; \rot s 2, 3 target narrator's friend), and \rot s that elaborate on additional social interpretation implicit in the situation (\rot~ 3: theft from a friend is cast as an act of betrayal). Effectively, \rot s represent evaluative judgments about a social situation in light of unspoken but accepted social norms.\footnote{Our definition of \rot s corresponds to the first of the two evaluative moral judgments defined in \citet{malle2014theory}.} Figure \ref{fig:figure2} shows three subsets of a situation's annotation to illustrate the perspectives RoTs capture.

\paragraph{Cultural Scope of this study}
We recognize that social norms are often culturally sensitive \cite{haidt1993affect,kagan1984nature} and judgments of morality and ethics concerning individuality, community and society do not always hold universally \cite{shweder1990defense}.
While some situations (e.g., ``punching someone'') might have similar levels of acceptability across a number of cultures, others might have drastically varied levels depending on the culture of its participants (e.g., ``kissing someone on the cheek as a greeting'').
As a starting point, our study focuses on the socio-normative judgments of English-speaking cultures represented within North America.
While we find some variation of judgments in our annotations (e.g., with respect to certain worker characteristics, see \S\ref{sup:worker-demographics}), extending this \formalism~to other countries and non-English speaking cultures remains a compelling area of future research.
\section{\datasetName}

We obtained 104k source situations from 4 text domains (\S\ref{sec:situations}), for which we elicited 292k \rot s from crowd workers (\S\ref{sec:rots}). We then define a structured annotation task where workers isolate the central action described by the \rot{} and provide a series of judgments about the \rot{} and the action (\S\ref{sec:rot-breakdown}). In total, we collect 365k structured annotations, performing multiple annotations per \rot{} for a subset of the \rot s to study the variance in annotations. Figure \ref{fig:datasetStatistics} illustrates our dataset statistics.

\begin{figure}[t]

\begin{center}
\includegraphics[width=0.99\linewidth]{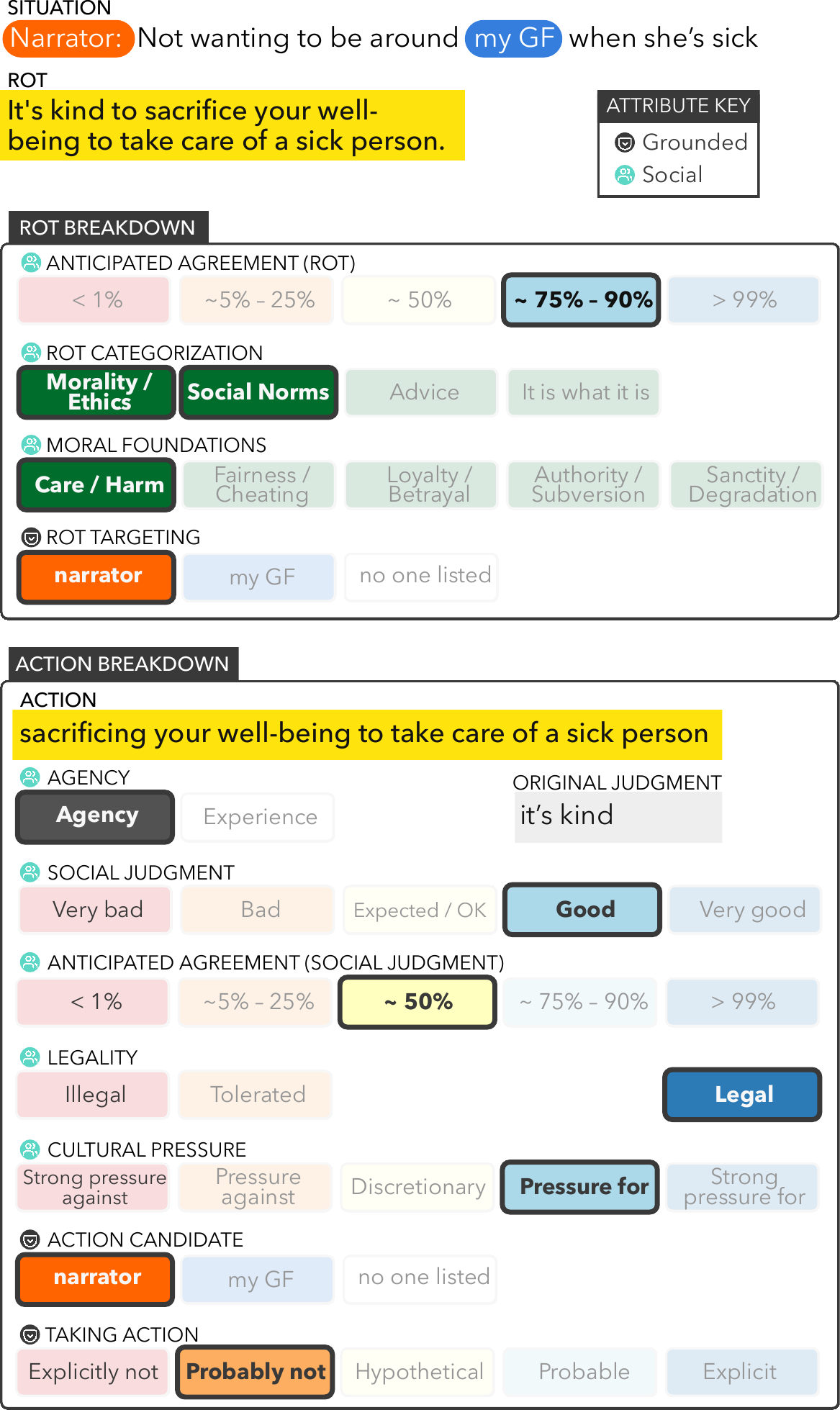}
\end{center}

\caption{
All attribute values for structured RoT annotations, with one complete example annotation filled in.
}

\label{fig:rot_breakdown_detail}
\end{figure}

\subsection{Situations}
\label{sec:situations}

\begin{figure*}[t!]

\begin{center}
\includegraphics[width=0.99\linewidth]{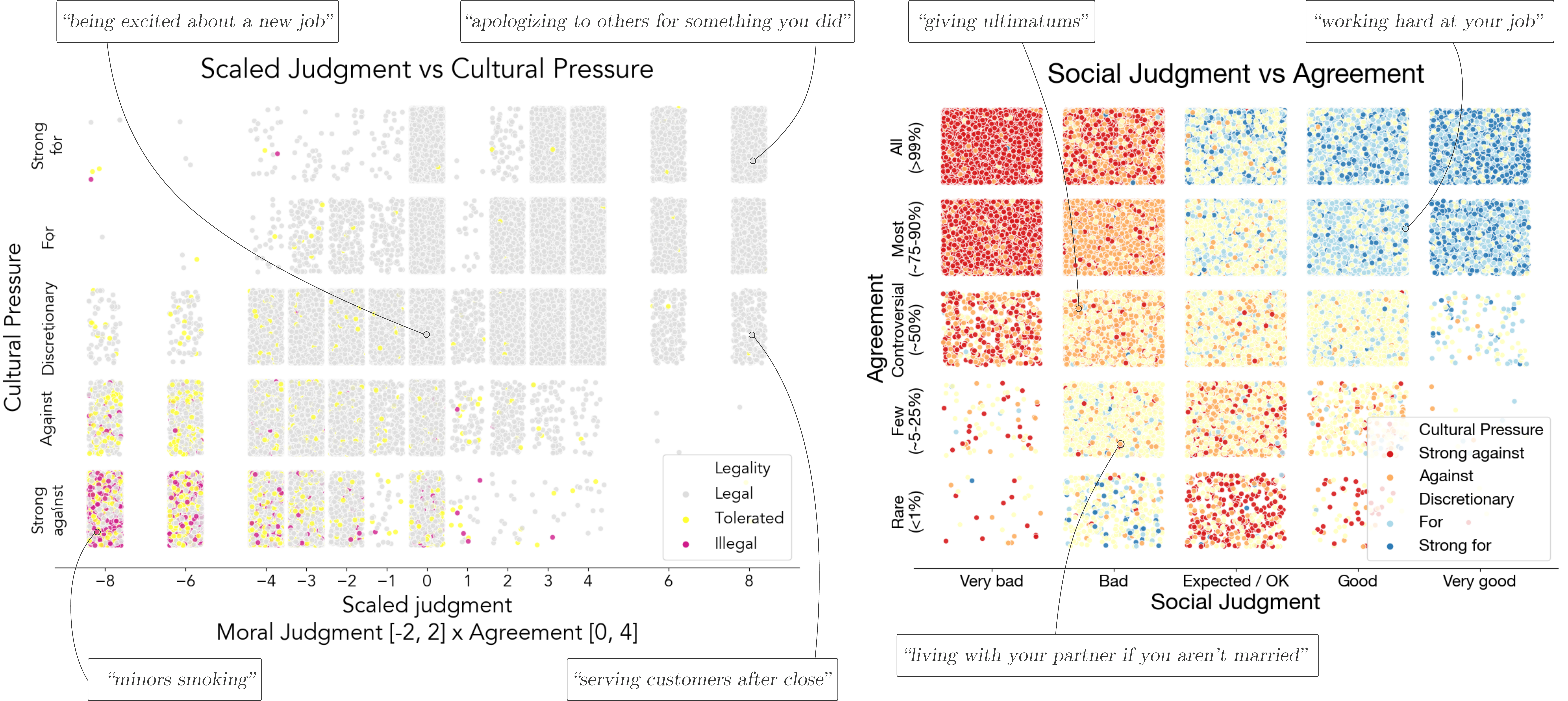}
\end{center}

\caption{
Plotting the distribution of \rot s in  \datasetNameNoDataset along axes of \textit{moral judgment, agreement, cultural pressure,} and \textit{legality.}
\textbf{Left:} Moral judgment is scaled with agreement (how commonly held the belief is) and plotted against cultural pressure.
Illegal activities 
fall in the bottom left: actions that are universally understood to be wrong
and people feel negative cultural pressure for.
\textbf{Right:} Moral judgment is plotted against agreement. 
Discretionary actions span a range of moral values (yellow ranging horizontally) and 
fringe beliefs often evoke strong negative cultural pressure even when morally neutral (bottom of plot).
}
\label{fig:datasetAnalysis}

\end{figure*}

We use a \textit{situation} to denote the one-sentence prompt given to a worker as the basis for writing \rot s. We gather a total of 104k real life situations from four domains: scraped titles of posts in the subreddits \texttt{r/confessions} (32k) and \texttt{r/amitheasshole} (\texttt{r/AITA}, 30k), which largely focus on moral quandaries and interpersonal conflicts; 30k sentences from the ROCStories corpus \cite[\texttt{rocstories},][]{mostafazadeh-etal-2016-corpus}; and scraped titles from the Dear Abby advice column archives\footnote{\href{https://www.uexpress.com/dearabby/archives}{https://www.uexpress.com/dearabby/archives}} (\texttt{dearabby}, 12k).\footnote{See Appendix~\ref{sup:domains} for further data preprocessing details.} 

\subsection{Rules-of-Thumb (\rot s)}
\label{sec:rots}

To collect \rot s, we provide workers with a situation as a prompt and them to write 1 -- 5 \rot s inspired by that situation. From the 104k situations, we elicit a total of 292k \rot s. Despite \rot s averaging just 10 words, we observe that 260k/292k \rot s are unique across the dataset.

For the development of \rot s, we instruct the workers to produce \rot s that \textit{explain the basics of social norms}, just as one would instruct a five-year-old child on the ABCs of acceptable 
conduct. \rot s are to be:
\begin{enumerate}
    \item \textbf{inspired by the situation}, to maintain a lower bound on relevance; 
    \item \textbf{self-contained}, to be understandable without additional explanation; and
    \item structured as \textbf{judgment} of acceptability (e.g., good/bad, (un)acceptable, okay) and an \textbf{action} that is assessed. 
\end{enumerate}
In order to encourage \rot{} diversity,
we also ask that an \rot{} 
should counterbalance
\textit{vagueness} 
against
\textit{specificity} so that 
\rot s generalize across multiple situations (e.g., \textit{``It is rude be selfish.''}) without being too specific (e.g., \textit{``It is rude not to share your mac'n'cheese with your younger brother.''}).
We also ask workers to write RoTs illustrating \textit{distinct ideas} and \textit{avoid trivial inversions} to prevent
low-information \rot s that rephrase the same idea or simply invert the judgement and action.

\paragraph{Character Identification.} We ask workers to identify phrases in each situation that refer to people.
For example, in a situation, like \textit{``\underline{My brother} chased after \underline{the Uber driver},''} workers mark the underlined spans. We collect three workers' spans, calling each span a \textit{character}. All characters identified become candidates for grounding \rot s and actions in the structured annotation. As such, we optimize for recall instead of precision by using the largest set of characters identified by any worker. We also include a \textit{\underline{narrator}} character by default. 

\subsection{\rot{} Breakdowns}
\label{sec:rot-breakdown}

We perform a structured annotation, which we term a \textit{breakdown}, on each \rot.
In an \rot{} breakdown, a worker isolates the underlying action contained in the \rot.
Then, they assign a series of categorical attributes to both the \rot{} and the action.
These categorical annotations allow for additional analyses and experiments relative to the text-only \rot s.

The attributes fall into two categories corresponding to the central annotation goals.
The first goal is to tightly \textit{ground} \rot s to their respective situations.
The second goal is to partition \textit{social} expectations using theoretically motivated categories. 

A subset of the attributes are labeled on the \rot{} (e.g., \textit{``It is expected that you report a crime''}), while others are on the action (e.g., \textit{``reporting a crime''}). Figure \ref{fig:rot_breakdown_detail} provides the complete set of labels available for an \rot{} breakdown.\footnote{Workers are given the choice to mark the \rot{} as confusing, vague, or low quality, and move on (2\% of \rot s).} 

\newcommand{\iconStdGrounded}{\includegraphics[width=8pt]{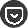}}
\newcommand{\iconStdSocial}{\includegraphics[width=8pt]{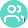}}

\vspace*{-1mm}
\paragraph{\iconStdGrounded~Grounding Attributes}
We call three attributes \textit{grounding attributes}.
Their goal is to ground the \rot{} and action to the situation and characters.
At the \rot-level, workers mark which character should heed the \rot~with the \textbf{\rot{}~Targeting} attribute.
At the action level, workers first pick the 
\textbf{action's best candidate}
character, for whom the action is most relevant.
However, since \rot s can identify actions that are both explicit and hypothetical in the situation,
we additionally annotate whether the candidate character is explicitly \textbf{taking the action} in the situation.

\vspace*{-1mm}
\paragraph{\iconStdSocial~Social Attributes} 
The second set of attributes characterize social expectations in an \rot.
The first two social attributes both label \textbf{anticipated agreement}.
For an \rot, this attribute asks how many people probably \textit{agree} with the \rot{} as stated.
At the action level, it asks what portion of people probably agree with the \textit{judgment} given the \textit{action}.

Four social attributes relate to the theoretical underpinnings of this work in \S\ref{sec:approach}.
An \rot-level attribute is the set of \textbf{Moral Foundations}, based on 
a well-known social psychology theory that outlines culturally innate moral reasoning \cite{haidt2012righteous}.
The action-level attributes \textbf{legality} and \textbf{cultural pressure} are designed to reflect the two-coarse-grained categories proposed by the Social Norms Theory \citep{kitts2008norms, perkins1986perceiving}.
Legality corresponds to prescriptive norms: what one ought to do.
Cultural pressure corresponds to descriptive norms: what one is socially influenced to do.
Finally, the \textbf{social judgment} aims to capture subjective moral judgment.
A base judgment of what is good or bad is thought to intrinsically motivate social norms  \citep{malle2014theory, haidt1993affect}.

The final two attributes provide a coarse categorization over \rot s and actions.
The \textbf{\rot{}~Category} attribute estimate distinctions between morality, social norms, and other kinds of advice.
This aims to separate moral directives from tips or general world knowledge (e.g., ``It is good to eat when you are hungry'').
The attribute \textbf{agency} is designed to let workers distinguish \rot s that involve agentive action from those that indicate an an experience (e.g., ``It is sad to lose a family member'').

\subsection{Analysis}

We briefly highlight three key aspects of our \formalism: social judgment, anticipated agreement, and cultural pressure. Figure \ref{fig:datasetAnalysis} shows two plots partitioning \rot s based on these three attributes (with legality also highlighted in the left plot (a)).

In the left plot (Figure \ref{fig:datasetAnalysis} (a)), the $x$-axis contains a new quantity, where social judgment ($\in [-2, 2]$) is multiplied by agreement ($\in [0, 4]$) to scale it.\footnote{Strict statisticians will note that plotting ordinal values numerically is an abuse of notation, much less scaling two values together.
We present these graphs for illustrative purposes to observe the stratification of our dataset, not to make quantitative claims.}
The result is that $x$ values range from universally-agreed bad actions (-8) to universally-agreed good actions (+8).
Intuitively, the bottom-left group shows illegal actions, which are both ``bad'' (left $x$) and people feel strong pressure not to do (bottom $y$).
The data are generally distributed in a line towards the top right, which are ``good'' (right $x$) actions that people feel strong pressure to do (top $y$).

However, the spread of the data in Figure \ref{fig:datasetAnalysis} (a) illustrates the difference between morality and cultural pressure.
There are a range of morally charged actions, but for which people don’t feel cultural pressure (the horizontal range in $x$ values across the central $y = \textit{Discretionary}$).
Conversely, we observe actions that are morally neutral, but for which people do feel cultural pressure (the vertical range in $y$ values along the middle $x=0$).

The right plot, Figure \ref{fig:datasetAnalysis} (b), shows social judgment against agreement, colored by cultural pressure.
At high levels of agreement (top of graph), cultural pressure (color) follows social judgment (horizontal changes in $x$ values).
However, for controversially-held judgments (lower $y$ values), we see a range of cultural pressure.
This includes morally good or bad actions that are still discretionary (middle $y$ values), as well as morally neutral actions for which people feel strong cultural pressure (lower $y$ values).

These plots illustrate two ways of stratifying actions along socially relevant dimensions. We anticipate considerable further dataset exploration remains.
\section{Model}
\label{sec:model}

We investigate neural models based on pre-trained language models for learning various sub-tasks derived from \datasetNameNoDataset.

\subsection{Training Objectives}

\begin{figure}[t]

\begin{center}
\includegraphics[width=0.99\linewidth]{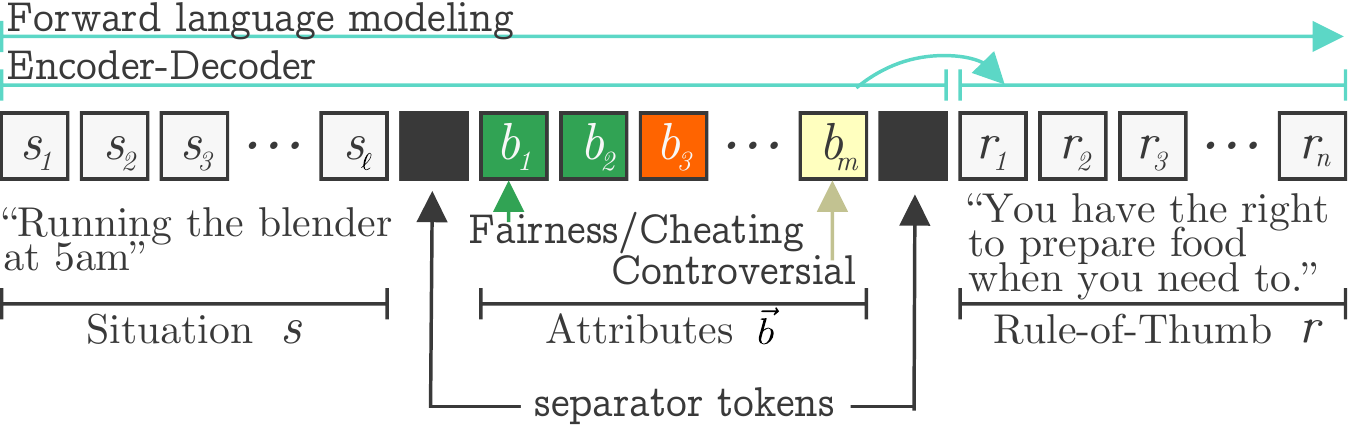}
\end{center}

\caption{
Illustration of modeling setup for the objective $p(r|s,\vec{b_r})$.
}
\label{fig:trainingDiagram}

\end{figure}

Our main modeling formulation is straightforward.
Given a situation ($s$), we wish to model the conditional distribution of \rot s ($r$), actions ($a$), and set of attributes from the breakdown ($\vec{b}$).
We can partition the attributes $\vec{b} = \{\vec{b_r}, \vec{b_a}\}$ into disjoint sets relevant to the \rot~ and action, and write
\begin{equation} \label{eq:decomposition}
p(r, a, \vec{b} | s) = \underbrace{p(a, \vec{b_a} | r, \vec{b_r}, s)}_{\textit{\scriptsize{action transcription}}} \times \underbrace{p(r, \vec{b_r} | s)}_{\textit{\scriptsize{\rot~ prediction}}}.
\end{equation}
\vspace{-3pt}

\setlength\tabcolsep{8pt}
\begin{table}[t]
\centering
\small
\begin{tabular}{@{}lll@{}}
\toprule
\multicolumn{2}{c}{\textit{Objective}} &                                   \\ \cmidrule(r){1-2}
\textbf{RoT} & \textbf{Action} & \textbf{Interpretation} \\ \midrule
$p(r|s)$ & $p(a|s)$ & Text-only generation                   \\ 
$p(\vec{b_r}|s)$ & $p(\vec{b_a}|s)$ & Attribute prediction \\
$p(r|s,\vec{b_r})$ & $p(a|s,\vec{b_a})$ & Controlled generation \\
$p(\vec{b_r}|s,r)$ & $p(\vec{b_a}|s,a)$ & Attribute labeling \\
$p(r,\vec{b_r}|s)$ & $p(a,\vec{b_a}|s)$ & Model choice generation \\ \bottomrule
\end{tabular}
\caption{Generative model objectives corresponding to the training setups we consider. Each model (\rot{} or action) is trained on all objectives simultaneously.
}
\label{tab:generative_model_setups}
\end{table}
\setlength\tabcolsep{6pt}

Equation \ref{eq:decomposition} allows us to model all components of interest given a situation $s$.
However, the \textit{action transcription} term is quite strongly conditioned, because actions are so closely related to their \rot s.
In this paper, we instead focus our study of actions on a more difficult distribution that conditions only on the situation:
\begin{equation} \label{eq:substitution}
 \underbrace{p(a, \vec{b_a} | r, \vec{b_r}, s)}_{\textit{\scriptsize{action transcription}}} \xrightarrow{\textit{\scriptsize{omit \rot}}} \underbrace{p(a, \vec{b_a} | s)}_{\textit{\scriptsize{action prediction}}}.
\end{equation}

We model both the \textit{\rot~ prediction} (Eq. \ref{eq:decomposition}) and \textit{action prediction} (Eq. \ref{eq:substitution}) distributions with conditional forward language modeling.
We tokenize all quantities ($s, r, a, \vec{b}$), creating unique tokens for each attribute value $b_i$, and concatenate them together in a canonical order to form strings $p(x_\text{out}|x_\text{in})$.
We then train to maximize the standard language modeling objective:
\begin{equation} \label{eq:linearization}
x = [x_\text{in};x_\text{out}], \quad p(x) = \prod_{i=1}^n{p(x_i|x_{<i})}.
\end{equation}

Both the \textit{\rot~ prediction} (Eq. \ref{eq:decomposition}) and \textit{action prediction} (Eq. \ref{eq:substitution}) distributions have similar forms $p(y,\vec{b_y}|s)$ for $y \in \{r, a\}$.
We take advantage of this symmetry to study variations of both distributions.
Inspired by recent work \cite{zellers2019neuralfakenews}, we construct permutations of our data that omit different fields while maintaining the canonical order.
Table \ref{tab:generative_model_setups} shows the setups that we consider, and Figure \ref{fig:trainingDiagram} illustrates an example objective.

We train each model (either RoT or action) on all relevant objectives in Table \ref{tab:generative_model_setups} (i.e., one of the columns).
Intuitively, this allows the model to condition on and generate a range of fields.\footnote{It is possible to remove the assumption that the situation is provided, which would allow the model to generate $s$ as well. We leave such experiments for future work.}
We can do this by simply treating each objective as defining a subset of the fields, as well as their ordering, for each data point.
Then, we combine and shuffle all objectives' views of the data.

\begin{table*}[ht]
\resizebox{\textwidth}{!}{%
\begin{tabular}{@{}lrrrrrrrrrrrrl@{}}
\toprule
 & \multicolumn{4}{c}{\textit{\textbf{$\rightarrow$ RoT}}}  & & \multicolumn{7}{c}{\textit{\textbf{$\rightarrow$ Action}}} &  \\ \cmidrule(r){2-5} \cmidrule(r){7-13}
 & \textbf{Category} & \textbf{Moral F.} & \textbf{Agree} & \textit{\textbf{Relevance}} & & \textbf{Agency} & \textbf{Judgment} & \textbf{Agree} & \textbf{Pressure} & \textbf{Legal} & \textbf{Taking} & \textit{\textbf{Relevance}} &   \\ \midrule
Random RoT          & 0.73 & 0.84 & 0.48 & 1.25 & & 0.90 & 0.57 & \textbf{0.55} & 0.53 & 0.80 & 0.04 & 1.22 &  \multirow{8}{*}{\rotatebox{270}{\small Model choice \enspace $p(y,\vec{b_y}|s)$}} \\
BERT-Score (Z et al., 2020)  & \textbf{0.76} & 0.83 & 0.48 & 2.00 & & 0.90 & \textbf{0.64} & 0.46 & \textbf{0.61} & 0.81 & 0.20 & 2.00 \\
GPT (R et al., 2018)             & 0.71 & 0.77 & 0.39 & 2.23 & & 0.82 & 0.40 & 0.36 & 0.32 & 0.76 & 0.15 & 2.25 \\
BART (L et al., 2019)            & 0.69 & 0.79 & \textbf{0.49} & 2.60 & & \textbf{0.91} & 0.55 & 0.54 & 0.46 & 0.80 & 0.18 & 2.52 \\
T5 (R et al., 2019)              & 0.62 & \textbf{0.85} & 0.42 & \textbf{2.78} & & 0.78 & 0.36 & 0.36 & 0.23 & 0.56 & 0.23 & \textbf{2.73} \\
GPT-2 Small (R et al., 2019)    & 0.62 & 0.79 & 0.34 & 2.03 & & 0.82 & 0.34 & 0.34 & 0.27 & 0.79 & 0.09 & 1.99 \\
GPT-2 XL - No pre-train      & 0.68 & 0.78 & 0.20 & 1.37 & & 0.81 & 0.37 & 0.30 & 0.33 & 0.79 & 0.06 & 1.29 \\
GPT-2 XL  & 0.75 & 0.84 & 0.42 & 2.53 & & \textbf{0.91} & 0.51 & 0.36 & 0.45 & \textbf{0.82} & \textbf{0.32} & 2.60 \\ \midrule
Random RoT          & 0.59 & 0.75 & \textbf{0.41} & 1.20 & & 0.84 & 0.27 & 0.28 & 0.21 & 0.74 & 0.01 & 1.19 & \multirow{8}{*}{\rotatebox{270}{\small Controlled \enspace $p(y|s,\vec{b_y})$}} \\
BERT-Score (Z et al., 2020)      & 0.66 & 0.78 & \textbf{0.41} & 2.00 & & 0.87 & 0.40 & \textbf{0.45} & 0.34 & \textbf{0.76} & 0.16 & 1.97 \\
GPT (R et al., 2018)             & 0.64 & 0.79 & 0.36 & 2.21 & & 0.83 & 0.46 & 0.36 & 0.38 & 0.74 & 0.17 & 2.26 \\
BART (L et al, 2019)            & 0.70 & \textbf{0.81} & 0.38 & 2.60 & & 0.84 & 0.47 & 0.42 & 0.41 & 0.73 & 0.20 & 2.44 \\
T5 (R et al., 2019)              & 0.66 & 0.80 & 0.40 & \textbf{2.77} & & 0.83 & 0.41 & 0.34 & 0.38 & 0.73 & 0.24 & \textbf{2.79} \\
GPT-2 Small (R et al., 2019)    & 0.64 & 0.78 & 0.30 & 2.10 & & 0.78 & 0.38 & 0.30 & 0.27 & 0.71 & 0.10 & 1.97 \\
GPT-2 XL - No pre-train      & 0.67 & 0.79 & 0.23 & 1.35 & & 0.83 & 0.36 & 0.32 & 0.26 & 0.73 & 0.04 & 1.33 \\
GPT-2 XL & \textbf{0.71} & 0.79 & 0.38 & 2.65 & & \textbf{0.90} & \textbf{0.51} & 0.38 & \textbf{0.42} & 0.74 & \textbf{0.28} & 2.54 \\
\bottomrule
\end{tabular}
}
\caption{Human evaluation results for conditionally generating \rot{}s and actions, either letting the models choose the attributes (top half), or providing the attributes as input constraints (bottom half). All columns are micro-F1 scores (0--1), except \textit{Relevance} (1--3). \textbf{Takeaway:} While state-of-the-art models are able to generate relevant \rot{}s and actions that generally follow constraints (moderately high scores in some columns), correctly conditioning on a complete set of attributes remains challenging (several columns show poor model performance in bottom half).
}
\label{tab:human_eval_generative_results}
\end{table*}
 
\subsection{Architectures}
We present results for the GPT and GPT-2 architectures \cite{gpt,gpt2}, as well as two encoder-decoder language models \cite[BART and T5,][]{lewis2019bart,raffel2019exploring}.
We train forward language models with loss over the entire sequence $x$, whereas encoder-decoder models only compute loss for the output sequence $x_\text{out}$.
Collectively, we term these architectures trained on our objectives the \modelName.

\section{Experiments and Results}

\subsection{Tasks}

While we train each model on all (RoT or action) objectives at once, we pick two particular objectives to asses the models.
The first is $p(y,\vec{b_y}|s)$ --- \textit{``model choice.''}
In this setting, each model is allowed to pick the most likely attributes $\vec{b_y}$ given a situation $s$, and generate an \rot~(or action) $y$ that adheres to those attributes.
This setup should be easier because a model is allowed to pick the conditions of its own generation ($\vec{b_y}$).

The second setting is $p(y|s,\vec{b_y})$ --- \textit{``conditional.''}
We provide models with a set of attributes $\vec{b_y}$ that they must follow when generating an \rot~(or action) $y$.
This presents a more challenging setup, because models cannot simply condition on the set of attributes that they find most likely. 
We select sets of attributes $\vec{b_y}$ provided by the human annotators for the situation $s$ to ensure models are not tasked with generating from impossible constraints.

\vspace*{-1mm}
\paragraph{Setup} We split our dataset into 80/10/10\% train/dev/test partitions by situation, such that each domain's situations are proportionally distributed. This guarantees previously unobserved dev and test situations. For all models we use top-$p$ decoding with $p=0.9$ \cite{Holtzman2020The}.

\vspace*{-1mm}
\paragraph{Baselines}
\label{sec:baselines}

We use a \textit{Random RoT} baseline to verify the dataset diversity (selections should have low relevance to test situations) and evaluation setup (RoTs and actions should still be internally consistent).
We also use a \textit{BERT-Score} \cite{Zhang2020BERTScore} retrieval baseline that finds the most similar training situation.
If attributes $\vec{b_y}$ are provided, the retriever picks the RoT (or action) from the retrieved situation with the most similar attributes.

\vspace*{-1mm}
\paragraph{Ablations} We report two model ablations. 
For \textit{-Small}, we finetune GPT-2 Small with the same general architecture.
For \textit{-No pretrain}, we randomly initialize the model's weights.\footnote{We omit the evaluation of an ``out-of-the-box GPT2-XL'' baseline (i.e. no fine-tuning) whose outputs predictably do not resemble \rot{}s or actions.}


\setlength\tabcolsep{2pt}
\begin{table}[t]
\centering
\scriptsize
\begin{tabular}{lrrr}
\textbf{Model} & \textbf{Ppl.} & \textbf{BLEU-4} & \textbf{Attr. $\mu$F1} \\ \toprule
\multicolumn{4}{l}{\textbf{$\rightarrow$ RoT}} \\ \midrule
GPT & 1.81 & 5.41 & 0.42 \\ 
Bart-large & 1.76 & 6.65 & 0.47 \\
T5-large & 1.94 & \textbf{10.79} & 0.34 \\
GPT-2 Small & 1.97 & 4.97 & 0.38 \\
GPT-2 XL - No fine-tune & - & 0.46 & 0.20 \\
GPT-2 XL - No pre-train \hspace{2em} & 2.54 & 4.39 & 0.42 \\
GPT-2 XL & 1.75 & 6.53 & \textbf{0.53} \\
\midrule
\multicolumn{4}{l}{\textbf{$\rightarrow$ Action}} \\ \midrule
GPT & 1.80 & 6.75 & 0.60 \\ 
BART-Large & 1.72 & 8.34 & 0.66 \\
T5-Large & 2.00 & \textbf{8.93} & 0.58 \\
GPT-2 Small & 1.94 & 6.62 & 0.56 \\
GPT-2 XL - No fine-tune & - & 0.25 & 0.52 \\
GPT-2 XL - No pre-train & 2.51 & 5.43 & 0.55 \\
GPT-2 XL & 1.73 & 7.98 & \textbf{0.68} \\ 
\bottomrule
\end{tabular}
\caption{
Test set performance by automatic metrics, including an attribute classifier. Perplexities are not comparable between encoder-decoder models (Bart and T5, loss on $x_\text{out}$ only) and other models (loss on full sequence $x$). \textbf{Takeaway:} Automatic metrics corroborate human evaluation results: while T5 is most adept at BLEU, GPT-2 XL more consistently adheres to attributes (Attr. $\mu$F1).
}
\label{tab:auto_metrics_generative_results}
\end{table}
\setlength\tabcolsep{6pt}


\subsection{Results}

\paragraph{Human Evaluation}
\label{sec:human_eval}

Table~\ref{tab:human_eval_generative_results} presents a human evaluation measuring how effective models are at generating \rot s and actions for both task settings.
While most columns measure attribute adherence, the \textit{Relevance} score is critical for distinguishing whether RoTs actually apply to the provided situation (e.g., see low scores for the \textit{Random RoT} baseline).
In both setups, T5's generations rank as most tightly relevant to the situation.
But in terms of correctly following attributes, GPT-2 is more consistent, especially in the \emph{controlled} task setup (lower; top scores on 5/9 attributes).
However, no model is able to achieve a high score on all columns in the bottom half of the table.
This indicates that fully constrained conditional generation may still present a significant challenge for current models.

\paragraph{Automatic Evaluation}
\label{sec:auto_eval}

We also provide automatic metrics of the generated outputs. We train attributes classifiers using RoBERTa \cite{Liu2019RoBERTaAR}, and use them to classify the model outputs.\footnote{BERT and BART performed worse across attributes.}

Table~\ref{tab:auto_metrics_generative_results} presents test set model performance on perplexity, BLEU \cite{papineni2002bleu}, and attribute micro-F1 classifier score. The automatic metrics are consistent with human evaluation. 
T5 is a strong generator overall, achieving the highest BLEU score and the highest \textit{relevance} score in \S\ref{sec:human_eval}. However, GPT-2 more consistently adheres to attributes, outperforming T5 in attribute $F_1$ with nearly 20 points gap for \rot{}s, and over 10 points for actions.

\section{Morality \& Political Bias}\label{sec:news-bias}

\begin{table}[t]
\small
    \centering
    \resizebox{\linewidth}{!}{%
    \begin{tabular}{@{}llr@{}lr@{}l@{}}
    \toprule
       &           & \multicolumn{2}{@{}c}{Left (-) or Right (+)} &     \multicolumn{2}{@{}c@{}}{Reliability} \\
    \toprule
    & Agreement &     -0.015 & $^{**}$ &        -0.008 & $^{*}$  \\ \midrule
    \multirow{4}{*}{\rotatebox{90}{ROT Cat.}} & Morality / Ethics &   -0.069 & $^{***}$ &      -0.022 & $^{***}$ \\
   & Social Norms &     0.019 & $^{***}$ &  -0.006 & $^{*}$ \\
   & It is what it is &    0.039 & $^{***}$ &        -0.007 & $^{**}$ \\
   & Advice &    0.031 & $^{***}$ &       0.033 & $^{***}$ \\ \midrule
    \multirow{5}{*}{\rotatebox{90}{Moral F.}}
   & Care / Harm &   -0.033 & $^{***}$ &      -0.016 & $^{***}$ \\        
   & Authority / Subversion &    \textit{n.s.} &  &       \textit{n.s.} &  \\
   & Fairness / Cheating &   -0.050 & $^{***}$ &       \textit{n.s.} &  \\
   & Loyalty / Betrayal &    0.026 & $^{***}$ &        -0.007 & $^{**}$ \\   
   & Sanctity / Degradation &      0.014 & $^{**}$ &      -0.017 & $^{***}$ \\
    \bottomrule
    \end{tabular}
    }
    \caption{
    Correlations between generated RoT attributes for headlines and the news source's political leaning (left: neg., right: pos.) and reliability (controlled for political leaning).
    Results shown are significant after Holm-correction for multiple comparisons {\small($p<0.001$: $^{***}$, $p<0.01$: $^{**}$, $p<0.05$: $^{*}$, $p>0.05$: \textit{n.s.})}.
    \textbf{Takeaway:} We see evidence that a model trained on the \datasetName can naturally uncover moral and topical leanings in news sources, mirroring results found in previous news studies.
    }
    \label{tab:news-bias}
\end{table}

To demonstrate a use case of our proposed \formalism, we analyze the social norms and expectations evoked in news headlines from news sources of various political leanings and trustworthiness, using the \modelName (GPT-2 XL).
Specifically, we generate ROTs and attributes for 50,000 news headlines randomly selected from \citet{Norregaard2019-bc}, a large corpus of political headlines from 2018 paired with news source
ratings of political leaning (5-point scale from left- to right-leaning) and factual reliability (5-point scale from least reliable to most reliable).\footnote{We use the MediaBias/FactCheck ratings: \url{https://mediabiasfactcheck.com}.} 

Table \ref{tab:news-bias} shows the correlations between \rot{} attributes and the political leaning and reliability of sources. 
Our results strongly corroborate 
findings by  \citet{Graham2009moralFoundationsLexicon}, showing that liberal headlines evoke more ``fairness'' and ``care,'' while right-leaning headlines evoke more ``sanctity'' and ``loyalty.''
Furthermore, in line with findings by \newcite{Volkova2017-xf}, more reliable news source tend to evoke more advice and less morality.

\section{Related Work}

Our \formalism{} heavily draws from works in descriptive ethics and social psychology,
but is also inspired by studies in social implicatures and cooperative principles in pragmatics \cite{kallia2004linguistic, grice1975logic} and the theories of situationally-rooted evocation of frames \cite{fillmore2001frame}.

Our work adds to the growing literature concerned with distilling reactions to situations \cite{vu2014acquiring,Ding2016AcquiringKO} as well as social and moral dynamics in language \cite{van-hee-etal-2015-detection}.
Commonly used for coarse-grained analyses of morality in text \cite{Fulgoni2016-by,Volkova2017-xf,Weber2018-hm}, \citet{Graham2009moralFoundationsLexicon} introduce the Moral Foundations lexicon, a dictionary of morality-evoking words \cite[later extended by][]{Rezapour2019EnhancedMoralF}. 

A recent line of work focused on representing social implications of everyday situations in free-form text in a knowledge graph \cite{rashkin2018event2mind,Sap2019ATOMICAA}.
Relatedly, \citet{sap2020socialbiasframes} introduce Social Bias Frames, a hybrid free-text and categorical formalism to reason about biased implications in language.
In contrast, our work formalizes a new type of reasoning around expectations of social norms evoked by situations.

Finally, concurrent works have developed rich and exciting resources studying similar phenomena.
\newcite{tay2020would} study \textit{Would you rather?} questions, and \newcite{acharya2020atlas} investigate ritual understanding across cultures. 
\newcite{hendrycks2020aligning} study ethical questions, attempting to assign a real-valued utility to scenarios across a range of ethical cateogires.
And \newcite{Lourie2020Scruples} define the challenge of predicting the \textit{r/AITA} task using the full posts.
In contrast to these studies, our work addresses norms by distilling cultural knowledge to a new conceptual level of Rules-of-Thumb and corresponding structural annotations.
\section{Conclusion}

We present \datasetNameNoDataset, an attempt at providing a \formalism~and resource around the study of grounded social, moral, and ethical norms.
Our experiments demonstrate preliminary success in generative modeling of structured \rot s, and corroborate findings of moral leaning in an extrinsic task.
Comprehensive modeling of social norms presents a promising challenge for NLP work in the future.

\section*{Acknowledgments}
The authors would like to thank Nicholas Lourie, Rowan Zellers, Chandra Bhagavatula, and Liwei Jiang.
This material is based upon work supported by the National Science Foundation Graduate Research Fellowship under Grant No. DGE1256082, and in part by NSF (IIS-1714566), DARPA CwC through ARO (W911NF15-1-0543), DARPA MCS program through NIWC Pacific (N66001-19-2-4031), and the Allen Institute for AI.

\bibliography{emnlp2020,anthology}
\bibliographystyle{acl_natbib}

\appendix
\section{Additional Dataset Details}
\subsection{Situations}

\paragraph{Domains}
\label{sup:domains}
We provide here a more thorough description how we collected situations from the four domains we consider. Figure \ref{fig:situation_examples} gives more example situations from each domain.

\begin{enumerate}
    \small
    
    \item \textbf{\texttt{r/amitheasshole}} (30k) --- The \textit{Am I the Asshole? (AITA)} subreddit.
    This posts of this subreddit pose moral quandries, such as \textit{``AITA for wanting to uninvite an (ex?)-friend from my wedding for shit-talking our marriage?''}
    We use the data from \newcite{Lourie2020Scruples}.
    They scrape the titles of posts, omitting the preamble (e.g., \textit{``AITA for''}), normalizing to present tense, and filtering out administrative posts.
    We do not use any annotations provided by that community (where other posters vote who had the moral high ground).
    
    \item \textbf{\texttt{r/confessions}} (32k) --- The \textit{Confessions} subreddit.
    This posts of this subreddit discuss personal stories, often with interpersonal conflicts, such as \textit{``I feel threatened by women prettier than me.''}
    As with \texttt{r/AITA}, we scrape only the titles of these posts.
    This subreddit contains a high volume of hateful or disturbing content; we attempt to filter the worst of this using keywords, and also allow annotators to mark dark or disturbing items.
    
    \item \textbf{\texttt{rocstories}} (30k) --- The ROCStories corpus from \citep{mostafazadeh-etal-2016-corpus}.
    ROCStories involve stories about everyday situations, and are generally less controversial than the other sources, e.g., \textit{``They weren't sure either so he started asking friends.''.}    
    We select a subset of the sentences from ROCStories which are likely to involve two character references based on POS tagging \cite{toutanova2003feature}, personal pronouns, and WordNet \cite{miller1995wordnet}.
    We then randomly sample to pick 30k sentences.
    
    \item \textbf{\texttt{dearabby}} (12k) --- Titles of the Dear Abby advice column.
    These titles are usually information dense summaries of interpersonal situations written in the style of news headlines, e.g., \textit{``Pushy Party Guests Make Themselves Too Much at Home.''}
    We scrape all of the titles found in the archives, and use heuristics to attempt to filter out all posts that do not match this style, such as announcements and holiday greetings.
    
\end{enumerate}

We attempt to balance the number of situations collected for each domain.
However, we are limited by the complete set of examples from \texttt{dearabby} (12k).

\paragraph{Additional Labels}

\begin{figure}[t!]

\begin{mdframed}

\scriptsize
\textbf{\texttt{[r/amitheasshole]}}
\begin{itemize}[noitemsep,topsep=0pt]
\renewcommand\labelitemi{--}
    \item telling my friend and her family to move out
    \item choosing to spend time with my friends or boyfriend rather than my family
    \item not wanting to hangout with sick girlfriend
    \item not wanting to do household chores
    \item banning my ex from my Spotify account
\end{itemize}

\smallskip

\textbf{\texttt{[r/confessions]}}
\begin{itemize}[noitemsep,topsep=0pt]
\renewcommand\labelitemi{--}
    \item My SO thinks I hate pickles, I like pickles but he LOVES pickles so I always pretend to hate them so he can have them.
    \item Best friend just got engaged.
    \item My girlfriend cheated and im cheating back on her
    \item I hate myself because I couldn't save my mother
    \item I'm scared of being a dad
\end{itemize}

\smallskip

\textbf{\texttt{[rocstories]}}
\begin{itemize}[noitemsep,topsep=0pt]
\renewcommand\labelitemi{--}
    \item Clark Ryder was proud of his job as a photojournalist.
    \item They had so many questions that I couldn't answer.
    \item Her husband surprised her on her birthday with plane tickets!
    \item She decided to wear slippers to protect her feet from Jason's toys.
    \item When he got to the assembled class he became very nervous.
\end{itemize}

\smallskip

\textbf{\texttt{[dearabby]}}
\begin{itemize}[noitemsep,topsep=0pt]
\renewcommand\labelitemi{--}
    \item Family of Six Tries Not to Be a Burden on Weekend Hosts
    \item Breakup Letter to Soldier Could Jeopardize Comrades in Arms
    \item Gentle Nudge Has Not Worked to Dislodge Mom From House
    \item Planning Helps Students Get Good Letter of Recommendation
    \item Man With Breast Cancer Experiences Extra Stress
\end{itemize}

\end{mdframed}

\caption{
Five randomly sampled situations from each of the four domains we consider.
}
\label{fig:situation_examples}

\end{figure}

We allow annotators to mark each situation with any of the following labels that apply.

\begin{itemize}
    \small
    \item \textbf{Unclear} The situation was too simple, vague, or confusing to understand what happened.
    \item \textbf{NSFW} The situation contains suggestive or adult content.
    \item \textbf{Dark / disturbing / controversial.}  The situation contained content that may make folks uncomfortable, like suicide, torture, or abuse.
\end{itemize}

Annotators may pass on writing RoTs for a situation marked with any of those boxes, or they may still choose to do so.
We keep all of the labels collected.
They are included in the dataset as additional fields.
For example, they could be used to omit certain training data to keep a model biased away from potentially controversial subjects.

\subsection{Character Identification}

Our goal during character identification is to find the most descriptive phrase referring to each unique non-narrator person in the passage exactly once.

The reason for this goal is that always having a single, best reference to each person in the situation enables more consistent grounding.

While this goal is relatively straightforward, we find many edge cases arise.
In cases where it is unclear if a person should be marked, our central criteria is \textbf{whether someone might write RoTs involving that person}.
If so, that person should be included so they are a candidate for grounding.
We found handling all of these edge cases complex enough to require human annotation instead of heuristics.
We provide here the character identification guidelines that we give to the crowd worker annotators, along with an example illustrating each one.

\paragraph{Character Identification Guidelines}

\begin{itemize}
\small
\renewcommand\labelitemi{--}

\item \textbf{Don't include the (first person) narrator.} For example, \textit{``I ate pizza''} would have no people highlighted.

\item \textbf{Only include people.} For example, \textit{``My horse George provides good conversation''} would have no people highlighted.

\item \textbf{Only highlight each person once.} For example, \textit{``I gave \underline{my brother} a hug, I like him, he's so nice''}, we would only include \underline{my brother}, not \textit{``him''} or \textit{``he.''}

\item \textbf{Highlight the most descriptive mention of a person.} For example, \textit{``I can't stand him, \underline{my brother} is so mean.''}, we would pick my brother even though it comes after \textit{``him.''}

\item \textbf{Include the full phrase referring to the person.} Include words like \textit{``a''}, \textit{``the''}, \textit{``my''}, and longer phrases. For example, \textit{``\underline{The strange guy} talked to \underline{my brother} and \underline{my oldest uncle},''} we would pick \underline{The strange guy}, \underline{my brother}, and \underline{my oldest uncle}, instead of just \textit{``guy''}, \textit{``brother''}, and \textit{``uncle.''}

\item \textbf{Don't include phrases where a generic person-looking word is used without referring to a particular person.} This often happens when describing a place or thing. For example, \textit{``I walked into the men's room,''} we would not pick anything, because \textit{``mens' room''} is a generic phrase. Similarly, we would not pick anything for, \textit{``I am a child,''} because \textit{``child''} is just used as a description. But for, \textit{``I walked into \underline{my brother}'s room''}, we would pick \underline{my brother}.

\item \textbf{Include people used to refer to someone.} For example, \textit{``\textbf{\underline{My brother}'s girlfriend} is so cool,''} we would pick both \underline{my brother} and \textbf{my brother's girlfriend}.

\item \textbf{Include pronouns (she, her, hers, etc.) \textit{if} they're the most specific word available.} For example, in a sentence like \textit{``I love \underline{him},''} we would pick \underline{him}. However, for a sentence like, \textit{``I love \underline{my brother}, I can always talk to him.''} we would instead pick \underline{my brother} because it's more specific.

\item \textbf{Include pronouns like \textit{``they''} and \textit{``them''}, also \textit{if} they're the most specific word available.} For example, if we had the sentence \textit{``\underline{They} went to the party.''} we would pick they. However, if we had the sentence \textit{``\underline{My friends} went to the party and they had a good time.''} we would instead pick \underline{My friends} since it is more specific.

\item \textbf{Include plural first person pronouns (us, we, etc.) once.} For example, in a sentence like \textit{``\underline{We} went to the park.''} we would pick we. Or for a sentence like \textit{``\underline{They} spent hours talking to \underline{us} and we had a good time.''} we would pick \underline{they} and \underline{us}.

\item \textbf{Include other groups of people like \textit{``her siblings,''} \textit{``their class,''} and \textit{``his team.''}} For example, in a sentence like \textit{``I talked to all of \textbf{\underline{his} uncles} for a while.''} we would pick both \underline{his} and \textbf{his uncles}.

\item \textbf{Include proper names of people that aren't the narrator.} For example, in a sentence like \textit{``\underline{Mary} chased \underline{John} at the park.''} we assume they are people (unless otherwise specified), and we would pick both \underline{Mary} and \underline{John}.

\item \textbf{Include people with titles like \textit{``the policeman''} and \textit{``the mailman.''}} For example, in the sentence \textit{``I chased \underline{the store clerk}.''} we would select \underline{the store clerk}.

\item \textbf{Include words like \textit{``someone''} and \textit{``everyone.''}} For example, in the sentence \textit{``I am going to dinner with \underline{someone}.''} we would select someone.

\end{itemize}

\subsection{Rules-of-Thumb (RoTs)}

\begin{figure*}[t!]

\begin{center}
\begin{mdframed}
\includegraphics[width=0.99\linewidth]{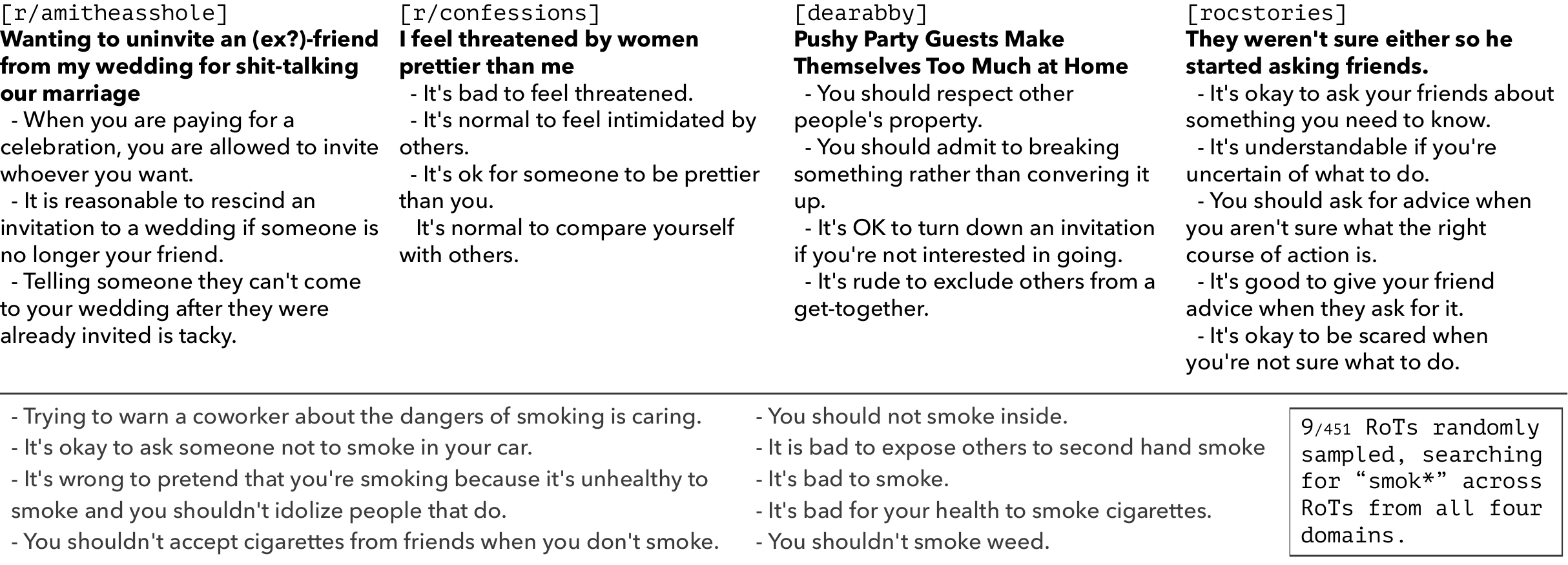}
\end{mdframed}
\end{center}

\caption{
\textbf{Top:} An example situation (bold) and corresponding RoTs (bullets) from each of the four domains we consider.
\textbf{Bottom:} Random sample of RoTs about smoking, found by searching for \textit{smok*} across the dataset.
}
\label{fig:rot_examples}

\end{figure*}

This section provides more information on how RoTs are written.
Figure \ref{fig:rot_examples} shows a sample of \rot s organized both by situation domain and topic.

As mentioned briefly in Section \ref{sec:rots} of the paper body, we present workers with a series of guidelines for how to write RoTs.
All RoT writing guidelines are in service of the goal that RoTs capture social, ethical, moral, and cultural norms.
Unlike the guidelines for character identification, which are largely syntactic, the guidelines for writing RoTs are semantic. This makes them more challenging both to define and check.

\definecolor{good}{HTML}{006D2C}
\definecolor{bad}{HTML}{A50F15}
\newcommand{\follows}{\textcolor{good}{\textbf{Follows:~}}}
\newcommand{\violates}{\textcolor{bad}{\textbf{Violates:~}}}
\newcommand{\why}{\textbf{Why:~}}

To motivate these guidelines, and to help readers intuitively characterize what RoTs are, we present the RoT writing guidelines here at greater length, annotated with examples and explanations. For each guideline (\textbf{in bold}), we provide an example situation (\textit{in italics}) along with candidate RoTs that \textcolor{bad}{\textbf{violate}} or \textcolor{good}{\textbf{follow}} the guideline.

\paragraph{RoT Writing Guidelines}

\begin{itemize}
    \small
    \renewcommand\labelitemi{--}

    \item \textbf{Explain the basics of good and bad behavior.} RoTs should describe cultural expectations, as if to a file-year-old child who doesn't yet know how the world works.
    \begin{itemize}[noitemsep,topsep=0pt]
        \item Example situation: \textit{Not wanting to take tests to apply for college}
        \item \violates ``Studies have shown people perform best on tests after sleeping at least seven hours''
        \item \follows ``It's normal to be stressed out by exams''
        \item \why This broad guideline attempts to distinguish RoTs from encyclopedic knowledge. Instead, RoTs should contain everyday, commonsense knowledge about social norms and expectations. 
    \end{itemize}

    \item \textbf{Judgment and action.} An RoT must comtain a judgment and an action.
    \begin{itemize}[noitemsep,topsep=0pt]
        \item Example situation: \textit{Telling my husband he shouldn't buy his dream boat}
        \item \violates ``Boats are expensive''
        \item \follows ``It's mean to squash someone's dreams''
        \item \follows ``People should be open to discussing big purchases with their spouses''
        \item \why Requiring an \textit{action} helps ensure RoTs are about things peoeple do. Requiring a \textit{judgment} pushes statements to contain some information about norms and expectations.
    \end{itemize}
    
    \item \textbf{Self-contained.} An RoT must be fully understandable on its own, without the situation it came from.
    \begin{itemize}[noitemsep,topsep=0pt]
        \item Example situation: \textit{Being angry at my sister for not attending our fathers funeral because of his criminal history.}
        \item \violates ``It makes them feel bad''        
        \item \violates ``The father caused emotional distress to his daughter and the narrator should not judge her actions too harshly.''
        \item \follows ``If someone commits serious crimes, it's OK for family to cut off contact with them.''        
        \item \why Without this requirement, RoTs would not naturally generalize to new situations, and would trend too specific. The would also could contain much less information, as much of the semantic content could be left in the situation and only referred to by the RoT.
    \end{itemize}
    
    \item \textbf{Inspired by situation.} An RoT should be inspired by the situation it came from.
    \begin{itemize}[noitemsep,topsep=0pt]
        \item Example situation: \textit{Wanting to uninvite a friend from my wedding.}
        \item \violates ``It's rude to point at people you don't know''
        \item \follows ``It's devastating to be excluded from a wedding you were invited to''        
        \item \why Maintaining a link between RoT and situation allows for grounding RoTs during the structured annotation. Furthermore, since a different worker will likely provides the structural annotation for an RoT, relevance to the source situation helps ensure the worker understands the RoT's context and implications.
    \end{itemize}
        
    \item \textbf{Balance Specificity and Vagueness.} An RoT should be inspired by, and relevant to, the provided situation. However, a rule-of-thumb should also give a general rule for how people behave in society, so should apply to more than just the given situation.
    \begin{itemize}[noitemsep,topsep=0pt]
        \item Example situation: \textit{Not tipping my cashier last Tuesday}
        \item \violates ``Not tipping a cashier last Tuesday is rude''
        \item \violates ``It's rude to be cheap''
        \item \follows ``It's usually OK not to tip cashiers in retail or grocery stores''
        \item \why This requirement can be the hardest to assess because of its subjectivity. RoTs that are too specific are usually slight modifications of the situation that include a judgment, and don't describe underlying expectations. RoTs that are too vague often do describe norms, but the link to the situation can be so distant as to be misleading. Good RoTs may be somewhat specific, but explain both the underlying norms at play, and apply to other situations.
    \end{itemize}

    \item \textbf{Distinct ideas.} When multiple RoTs are provided for a situation, each should contain a distinct idea. This includes inversions of the same idea.
    \begin{itemize}[noitemsep,topsep=0pt]
        \item Example situation: \textit{Never taking out the trash}
        \item \violates ``It's irresponsible to avoid the chores you are assigned'' \textit{with} ``It's bad to not do chores you're supposed to do''
        \item \violates ``It's irresponsible to avoid the chores you are assigned'' \textit{with} ``It's responsible to do the chores you are assigned''        
        \item \follows ``It's irresponsible to avoid the chores you are assigned'' \textit{with} ``You're generally expected to have a clean home''        
        \item \why This requirement is to prevent merely collecting paraphrases of the same RoT. Furthermore, we assume that inversions are usually trivial semantic mutations, so they are also not worth collecting at scale.
    \end{itemize}

\end{itemize}

\subsection{RoT Breakdowns}

In this section, we provide more information about the structural RoT annotations, which we call ``RoT Breakdowns.''
In particular, we illustrate the potential values for each attribute with an example.

\newcommand{\iconNegNeg}{\includegraphics[width=7pt]{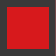}}
\newcommand{\iconNeg}{\includegraphics[width=7pt]{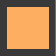}}
\newcommand{\iconMid}{\includegraphics[width=7pt]{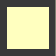}}
\newcommand{\iconPos}{\includegraphics[width=7pt]{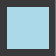}}
\newcommand{\iconPosPos}{\includegraphics[width=7pt]{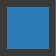}}

\newcommand{\iconGrounded}{\includegraphics[width=7pt]{icon/icon_grounded.pdf}}
\newcommand{\iconSocial}{\includegraphics[width=7pt]{icon/icon_social.pdf}}

\newcommand{\typeSocial}{\iconSocial~Social}
\newcommand{\typeGrounded}{\iconGrounded~Grounded}


\subsubsection{RoT Categorization}

RoT categories are originally designed to distinguish more desired annotation topics (morality/ethics, social norms) from less desired ones (advice and ``it is what it is'' statements). 
RoT categories are not mutually-exclusive, and the lines are not always clear.
While we use all data regardless of RoT category in this paper's experiments, future work using this dataset may consider filtering based on RoT category.
Annotators are allowed to select from none through all, but are encouraged to select the 1 -- 2 best.

\smallskip

\noindent {\small
\setlength{\extrarowheight}{0.3em}
\begin{tabular}{@{}p{0.16\linewidth}p{0.75\linewidth}}
\multicolumn{2}{@{}l}{\textbf{Information}} \\ \toprule
\textbf{Type} & \typeSocial \\
\textbf{Annotated} & RoT \\
\textbf{Prompt} &  \textit{What category best fits this RoT?}\\
\end{tabular}
}

\smallskip

\noindent {\footnotesize
\setlength{\extrarowheight}{0.4em}
\begin{tabular}{@{}p{0.19\linewidth}p{0.30\linewidth}p{0.30\linewidth}}
\multicolumn{2}{@{}l}{\textbf{Values}} \\ \toprule
\textbf{Label} & \textbf{Description} & \textbf{Example} \\ \midrule
Morality/ Ethics & Rules or guiding principles of right or wrong conduct & \textit{It's wrong to turn your back on your friends who need your help.} \\
Social Norms & Standards of appropriate behavior in a society; emphasizes social conventions & \textit{It’s good to shake hands with your opponent even if you lost.} \\
Advice & Prudent behaviors that are aimed at improving one’s life & \textit{It's good to take medicine your doctor prescribes.} \\
It is what it is & Describes how things are; avoids ethics, morality or social norms & \textit{It is nice to be tall.} \\
\end{tabular}
}

\smallskip


\subsubsection{Moral Foundations}

To simplify the annotation, we label \textit{axes} of moral foundations.
For example, \textit{Care/Harm} are annotated together, rather than as \textit{Care} and \textit{Harm} separately.
Other social attributes, such as \textit{social judgment}, attempt to explicitly capture the goodness or badness of the act.
Additionally, we omit the sixth moral foundation, \textit{Liberty/Oppression}, which was later added to the formalism after the first five.
Some examples are drawn from literature on Moral Foundations \cite{haidt2012righteous}.
Note that these labels are not mutually exclusive, and annotators may choose from none through all.

\smallskip

\noindent {\small
\setlength{\extrarowheight}{0.3em}
\begin{tabular}{@{}p{0.16\linewidth}p{0.75\linewidth}}
\multicolumn{2}{@{}l}{\textbf{Information}} \\ \toprule
\textbf{Type} & \typeSocial \\
\textbf{Annotated} & RoT \\
\textbf{Prompt} &  \textit{Which moral foundations best fit this RoT?}\\
\end{tabular}
}

\smallskip

\noindent {\footnotesize
\setlength{\extrarowheight}{0.4em}
\begin{tabular}{@{}p{0.19\linewidth}p{0.30\linewidth}p{0.30\linewidth}}
\multicolumn{2}{@{}l}{\textbf{Values}} \\ \toprule
\textbf{Label} & \textbf{Description} & \textbf{Example} \\ \midrule
Care/Harm & Preventing or inflicting pain or suffering & \textit{It’s mean to tell someone they aren't attractive as someone else.} \\
Fairness/ Cheating & Notions of equity, justice, and rights & \textit{It's wrong to copy someone else's answers in an exam.} \\
Loyalty/ Betrayal & Obligations or concerns for group, family, and nation. & \textit{It’s rude for a cheerleader to boo their own team.} \\
Authority/ Subversion & Submission and deference to traditions or legitimate authority. & \textit{It's rude to walk away from your boss while they are talking to you.} \\
Sanctity/ Degradation & Abhorrence for disgusting things or actions; purity & \textit{It’s disgusting to pee into a public pool full of people.} \\
\end{tabular}
}

\smallskip

\subsubsection{Action and Judgment}

Each RoT should contain and action and a judgment.
An essential step in the RoT brekadown is to identify the action so that it may be annotated with action attributes.
We have annotators write both the action and the judgment, conjugating to the action to gerund tense, and rewriting when needed.

If the annotator is unable to clearly separate and action and a judgment from the RoT, they can simply mark the RoT as unclear and move on.

We provide here a few examples of RoTs along with corresponding actions and judgments.
Some are straightforward, while others require more substantial rewriting.

\begin{itemize}
    \small
    \renewcommand\labelitemi{--}

    \item \textbf{RoT:} It's good to care for the elderly.
    \begin{itemize}[noitemsep,topsep=0pt]
        \item \textbf{Action:} caring for the elderly
        \item \textbf{Judgment:} it's good
    \end{itemize}
    
    \item \textbf{RoT:} It's okay to be angry if your friend talks to someone you used to date
    \begin{itemize}[noitemsep,topsep=0pt]
        \item \textbf{Action:} being angry when your friend talks to someone you used to date
        \item \textbf{Judgment:} it's okay
    \end{itemize}

    \item \textbf{RoT:} The holidays are expected to be especially difficult for those that are grieving.
    \begin{itemize}[noitemsep,topsep=0pt]
        \item \textbf{Action:} finding the holidays difficult while grieving
        \item \textbf{Judgment:} expected
    \end{itemize}
    
\end{itemize}


\subsubsection{Agency}

It can be challenging to distinguish \textit{agency} from \textit{experience} in cases where the action involves thinking thoughts or feeling emotions.
We provide the following additional examples to workers for these cases, and allow their discretion during the annotation:

\begin{itemize}[noitemsep,topsep=0pt]
    \small
    \renewcommand\labelitemi{--}
    \item \textbf{Experience:} Feeling upset when someone calls you a name
    \item \textbf{Agency:} Being mad for two days when someone calls you a name
    \item \textbf{Agency:} Taking revenge on someone for calling you a name
\end{itemize}

\smallskip

\noindent {\small
\setlength{\extrarowheight}{0.3em}
\begin{tabular}{@{}p{0.16\linewidth}p{0.75\linewidth}}
\multicolumn{2}{@{}l}{\textbf{Information}} \\ \toprule
\textbf{Type} & \typeSocial \\
\textbf{Annotated} & Action \\
\textbf{Prompt} &\textit{Is the action \underline{ \$action } something you do or control, or is it something you experience?}\\
\end{tabular}
}

\smallskip

\noindent {\footnotesize
\setlength{\extrarowheight}{0.4em}
\begin{tabular}{@{}p{0.17\linewidth}p{0.31\linewidth}p{0.31\linewidth}}
\multicolumn{2}{@{}l}{\textbf{Values}} \\ \toprule
\textbf{Label} & \textbf{Description} & \textbf{Example} \\ \midrule
Agency & You can do this & \textit{Doing the dishes} \\
Experience & This happens to you; you have no control & \textit{Being stuck in an earthquake} \\
\end{tabular}
}

\smallskip


\subsubsection{Social Judgment}

For social judgment, we ask workers to transcribe the intent of RoT's original judgment, rather than pick their own.
Then, workers can mark their disagreement through their annotation of the \textit{anticipated agreement} attribute.
This allows us to faithfully capture controversial opinions, while also marking them as controversial.\\

\smallskip

\noindent {\small
\setlength{\extrarowheight}{0.3em}
\begin{tabular}{@{}p{0.16\linewidth}p{0.75\linewidth}}
\multicolumn{2}{@{}l}{\textbf{Information}} \\ \toprule
\textbf{Type} & \typeSocial \\
\textbf{Annotated} & Action \\
\textbf{Prompt} &\textit{Which best matches the RoT's original judgment (\underline{ \$judgmentSpan }) of \underline{ \$action }?}\\
\end{tabular}
}

\smallskip

\noindent {\footnotesize
\setlength{\extrarowheight}{0.4em}
\begin{tabular}{@{}p{0.01\linewidth}p{0.17\linewidth}p{0.31\linewidth}p{0.31\linewidth}}
\multicolumn{2}{@{}l}{\textbf{Values}} \\ \toprule
 & \textbf{Label} & \textbf{Description} & \textbf{Example} \\ \midrule
\iconNegNeg & Very bad & Spiteful, malicious, or insulting & \textit{Slashing tires; Destroying someone’s phone} \\
\iconNeg & Bad & Frowned upon, rude, unkind, bad & \textit{Littering; saying something mean to a friend} \\
\iconMid & Expected / OK & Normal; generally expected by society & \textit{Wearing clothes; showing up to work on time} \\
\iconPos & Good & Kind of considerate & \textit{Driving a friend to the airport} \\
\iconPosPos & Very good & Generous, self-sacrificing, perhaps even heroic & \textit{Buying groceries for a financially struggling neighbor} \\
\end{tabular}
}

\smallskip


\subsubsection{Anticipated Agreement}

Anticipated agreement is theoretically continuously valued, compared to other attributes that are more naturally ordinal.
Inspired by \newcite{Wang2018ModelingSP}, we bin this attribute into intuitive ranges, primarily to reduce cognitive load during annotation.

\smallskip

\noindent {\small
\setlength{\extrarowheight}{0.3em}
\begin{tabular}{@{}p{0.16\linewidth}p{0.75\linewidth}}
\multicolumn{2}{@{}l}{\textbf{Information}} \\ \toprule
\textbf{Type} & \typeSocial \\
\textbf{Annotated} & RoT, Action \\
\textbf{Prompts} & \textbf{RoT:} \textit{What portion of people probably agree that \underline{ \$RoT }?}\\
                 & \textbf{Action:} \textit{What portion of people probably agree that \underline{ \$action } is \underline{ \$socialJudgment }?}\\
\end{tabular}
}

\smallskip

\noindent {\footnotesize
\setlength{\extrarowheight}{0.4em}
\begin{tabular}{@{}p{0.01\linewidth}p{0.19\linewidth}p{0.30\linewidth}p{0.30\linewidth}}
\multicolumn{2}{@{}l}{\textbf{Values}} \\ \toprule
 & \textbf{Label} & \textbf{Description} & \textbf{Example} \\ \midrule
\iconNegNeg & $< 1$\% & Almost no one & \textit{Murdering people is great} \\
\iconNeg & $\sim5$--$25$\% & People occasionally think this. & \textit{Ghosts exist} \\
\iconMid & $\sim50$\% & Controversial (people naturally disagree) & \textit{It's okay to have an abortion} \\
\iconPos & $\sim75$--$90$\% & Common belief & \textit{It's okay to try again when you fail} \\
\iconPosPos & $> 90$\% & Universally true & \textit{It's bad to deliberately hurt your pets} \\
\end{tabular}
}

\smallskip


\subsubsection{Legality}

Annotations of legality are usually straightforward.
Whether an action is illegal but tolerated can vary depending on location and knowledge of the law.\\

\smallskip

\noindent {\small
\setlength{\extrarowheight}{0.3em}
\begin{tabular}{@{}p{0.16\linewidth}p{0.75\linewidth}}
\multicolumn{2}{@{}l}{\textbf{Information}} \\ \toprule
\textbf{Type} & \typeSocial \\
\textbf{Annotated} & Action \\
\textbf{Prompt} & \textit{Where you live, how legal is the action \underline{ \$action }?}\\
\end{tabular}
}

\smallskip

\noindent {\footnotesize
\setlength{\extrarowheight}{0.4em}
\begin{tabular}{@{}p{0.01\linewidth}p{0.19\linewidth}p{0.30\linewidth}p{0.30\linewidth}}
\multicolumn{2}{@{}l}{\textbf{Values}} \\ \toprule
 & \textbf{Label} & \textbf{Description} & \textbf{Example} \\ \midrule
\iconNegNeg & Illegal & Legal consequences if caught  & \textit{Theft; murder} \\
\iconNeg & Depends/ Tolerated & Generally ``illegal'', but often unenforced depending on circumstances & \textit{Using a cellphone while driving} \\
\iconPosPos & Legal & Not illegal & \textit{Coughing without covering one’s mouth} \\
\end{tabular}
}

\smallskip


\subsubsection{Cultural Pressure}

We provide instructions that cultural pressure could come from one's family, friends, community, culture, or society at large.
We ask annotators to evaluate cultural pressure according to their own feelings.\\

\smallskip

\noindent {\small
\setlength{\extrarowheight}{0.3em}
\begin{tabular}{@{}p{0.16\linewidth}p{0.75\linewidth}}
\multicolumn{2}{@{}l}{\textbf{Information}} \\ \toprule
\textbf{Type} & \typeSocial \\
\textbf{Annotated} & Action \\
\textbf{Prompt} & \textit{How much cultural pressure do you (or those you know) feel about \underline{ \$action }?}\\
\end{tabular}
}

\smallskip

\noindent {\footnotesize
\setlength{\extrarowheight}{0.4em}
\begin{tabular}{@{}p{0.01\linewidth}p{0.19\linewidth}p{0.30\linewidth}p{0.30\linewidth}}
\multicolumn{2}{@{}l}{\textbf{Values}} \\ \toprule
 & \textbf{Label} & \textbf{Description} & \textbf{Example} \\ \midrule
\iconNegNeg & Strong pressure against & Culture frowns upon this action  & \textit{Intentionally harming an animal} \\
\iconNeg & Pressure against & Culture generally discourages this action & \textit{Spending money on jewelry if you can't afford it} \\
\iconMid & Discretionary & Culture has little or nothing to say about this action & \textit{Choosing to read before bed} \\
\iconPos & Pressure for & Culture generally encourages this action & \textit{Being honest with people} \\
\iconPosPos & Strong pressure for & Culture strongly promotes this action & \textit{Wearing clothes in public} \\
\end{tabular}
}

\smallskip


\subsubsection{Taking Action}

RoTs are written for a range of both hypothetical and actual actions related to the provided situation.
Furthermore, sometimes the action is one that is explicitly not happening.
This attribute labels how likely it is that the action is being taken by the relevant character.
\textit{Note: a subset of the \texttt{r/AITA} annotations were performed before the ``probably not'' label was introduced; for those, ``hypothetical'' is marked instead.}\\

\smallskip

\noindent {\small
\setlength{\extrarowheight}{0.3em}
\begin{tabular}{@{}p{0.16\linewidth}p{0.75\linewidth}}
\multicolumn{2}{@{}l}{\textbf{Information}} \\ \toprule
\textbf{Type} & \typeGrounded \\
\textbf{Annotated} & Action \\
\textbf{Prompt} & \textit{Is \underline{ \$candidateCharacter } explicitly doing the action \underline{ \$action }? Or is it the action \textbf{might} happen?}\\
\end{tabular}
}

\smallskip

The upcoming examples use \textit{narrator} and the following situation for context: \textit{Not tipping the bartender at the club.}\\

\smallskip

\noindent {\footnotesize
\setlength{\extrarowheight}{0.4em}
\begin{tabular}{@{}p{0.01\linewidth}p{0.19\linewidth}p{0.30\linewidth}p{0.30\linewidth}}
\multicolumn{2}{@{}l}{\textbf{Values}} \\ \toprule
 & \textbf{Label} & \textbf{Description} & \textbf{Example} \\ \midrule
\iconNegNeg & Explicitly not & It's explicitly written that they don't do this  & \textit{Tipping the bartender} \\
\iconNeg & Probably not & Most likely not; they probably don't do this & \textit{Enjoying the drinks} \\
\iconMid & Hypothetical & We can't say / no evidence & \textit{Going clubbing every day} \\
\iconPos & Probable & Most likely; hints are written & \textit{Paying for drinks} \\
\iconPosPos & Explicit & It's written in the situation & \textit{Going to the club} \\
\end{tabular}
}

\smallskip


\subsection{Crowdsourcing}

Workers undergo an extensive vetting process before working on \rot s.
This includes a paid qualification (qual) with a quiz on each of the guidelines and a manual review of sample \rot s.
Workers then pass the qual move to a staging pool where they can work on a small number of situations, and all of their \rot s are manually reviewed for adherence to the guidelines. 
After graduating from the staging pool, workers enter the main group of \rot~ writers and annotators.
For every batch of data, we perform spot checks on the \rot s written and annotated by the main group, as well as send feedback to all of the workers answering any questions we receive.
We continuously update the instructions with clarifications, new examples, and answers to questions.

\subsection{Annotator Demographics}\label{sup:worker-demographics}
With an extensive qualification process, 137 workers participated in our tasks.
Of those, 55\% were women and 45\% men.
89\% of workers identified as white, 7\% as Black.
39\% were in the 30-39 age range, 27\% in the 21-29 and 19\% in the 40-49 age ranges.
A majority (53\%) of workers were single, and 35\% were married.
47\% of workers considered themselves as middle class, and 41\% working class.
In terms of education level, 44\% had a bachelor's degree, 36\% some college experience or an associates degree.
Two-thirds (63\%) of workers had no children, and most lived in a single (25\%) or two-person (31\%) household.
Half (48\%) our workers lived in a suburban setting, the remaining half was evenly split between rural and urban.
Almost all (94\%) of our workers had spent 10 or more years in the U.S.

\subsection{Demographics and Annotations}\label{sup:analyses-demographics}

\begin{table*}[t]
\centering
\begin{tabular}{@{}lr@{}lr@{}lr@{}lr@{}l@{}}
\toprule
 &   \multicolumn{2}{p{2cm}}{RoT Agreement}   &        \multicolumn{2}{p{2cm}}{Action Agreement}  & \multicolumn{2}{p{2cm}}{Cultural Pressure} &   \multicolumn{2}{p{2cm}}{Social Judgment}  \\
\midrule
Gender (M: 0, F: 1)   &  0.070 & $^{***}$ &       0.104 & $^{***}$ &   \textit{n.s.} &  &         \textit{n.s.} &  \\
Urbanness             &  0.065 & $^{***}$ &       0.085 & $^{***}$ &   \textit{n.s.} &  &         \textit{n.s.} &  \\
Education             &   0.022 & $^{**}$ &       0.037 & $^{***}$ &   \textit{n.s.} &  &         0.025 & $^{***}$ \\
Politics (rep: 0, dem: 1)              &  0.052 & $^{***}$ &       0.075 & $^{***}$ &    0.023 & $^{**}$ &         \textit{n.s.} &  \\
Household size           &  0.059 & $^{***}$ &       0.080 & $^{***}$ &   \textit{n.s.} &  &         \textit{n.s.} &  \\
Social class           &  \textit{n.s.} &  &       \textit{n.s.} &  &   \textit{n.s.} &  &         \textit{n.s.} &  \\
Income                &   -0.027 & $^{*}$ &       \textit{n.s.} &  &   \textit{n.s.} &  &         \textit{n.s.} &  \\
Age                  &  \textit{n.s.} &  &       \textit{n.s.} &  &   \textit{n.s.} &  &         \textit{n.s.} &  \\

\bottomrule
\end{tabular}
\caption{Correlations between worker demographics and categorical RoT annotations, Bonferroni corrected for multiple comparisons ($p<0.0001$: $^{***}$, $p<0.001$: $^{**}$, $p<0.01$: $^{*}$).}
\label{tab:analyses-demographics}
\end{table*}

We analyze the demographic variation in \rot~ and action annotations, using a set of 400 \rot s that were annotated by 50 workers each.
In addition to the demographic variables described in \S\ref{sup:worker-demographics}, we also consider the political leaning of the state in which the worker resides (self-reported), by assigning each state a value based on the state-level voting patterns in the last four national elections (yielding five-point scale from 100\% republican to 100\% democratic).

For our analyses, we run a generalized linear model regressing the RoT categories
on all $z$-scored demographic variables, and report the $\beta$ coefficients from that model.
In our action moral judgment analyses, we control for actions; for action agreement, we control for the action and the moral judgement; for the RoT agreement and action pressure, we control for individual RoTs.
Our results for categorical RoT annotations are shown in Table \ref{tab:analyses-demographics}.

\paragraph{Agreement (RoT and Action)}
The projection of how many people agree with the judgement is correlated with various demographic characteristics.
Specifically, judgments of actions, being a woman and living in an urban setting was most strongly correlated with ascribing high agreement to the judgment.
Other associations include higher education, household size, and inferred political leaning based on state of residency.

For \rot~ agreement, we find similar but weaker associations. Additionally, we find a small correlation between income and social class and ascribing higher agreement.

\paragraph{Cultural Pressure}
The only variable correlated with feeling culturally pressured is the political leaning of the state where workers are located, though the effect is small.

\paragraph{Social Judgment}
Similar to action agreement.
Effects are somewhat weaker, but workers being women, highly educated, or younger are associated with selecting higher (better) judgment to actions.

\section{Experimental details}

\paragraph{Generative Models} We use the Transformers package \cite{Wolf2019HuggingFacesTS} to implement our models. We train all the models for a single epoch with a batch size of 64, with the random seed 42. Each input and output sequence is prefixed with a special token indicating its type (e.g. \texttt{[attrs], [rot], [action]}). We also define a special token for each attribute value (e.g. \texttt{$<$morality-ethics$>$, $<$bad$>$, $<$all$>$, $<$against$>$}). We initialize the special token embeddings with the embedding of their corresponding words, taking the average for multiword expressions. For example, $\vec{v}_{<\text{bad}>} = \vec{v}_{\text{bad}}, \vec{v}_{<\text{morality-ethics}>} = (\vec{v}_{\text{morality}} + \vec{v}_{\text{ethics}})/2$. 

\end{document}